\documentclass{article}

\PassOptionsToPackage{numbers,compress}{natbib}
\usepackage[preprint]{neurips_2026}

\usepackage[utf8]{inputenc}
\usepackage[T1]{fontenc}
\usepackage{hyperref}
\usepackage{url}
\usepackage{booktabs}
\usepackage{amsmath}
\usepackage{amssymb}
\usepackage{amsfonts}
\usepackage{nicefrac}
\usepackage{microtype}
\usepackage{xcolor}
\usepackage{graphicx}

\usepackage[ruled]{algorithm2e}

\SetAlFnt{\small}
\SetAlCapFnt{\small}
\SetAlCapNameFnt{\small}
\SetAlCapHSkip{0pt}

\usepackage{xspace}
\usepackage{adjustbox}
\usepackage{bm}
\usepackage{multirow}
\usepackage{multicol}
\usepackage{colortbl}
\usepackage{pifont}
\usepackage{placeins}
\usepackage{flafter}
\usepackage{wrapfig}
\usepackage{subfig}

\usepackage[capitalize]{cleveref}
\Crefname{section}{Section}{Sections}
\crefname{section}{Sec.}{Secs.}
\Crefname{align}{Equation}{Equations}
\crefname{align}{Eq.}{Eqs.}
\Crefname{equation}{Equation}{Equations}
\crefname{equation}{Eq.}{Eqs.}
\Crefname{figure}{Figure}{Figures}
\crefname{figure}{Fig.}{Figs.}
\Crefname{table}{Table}{Tables}
\crefname{table}{Tab.}{Tabs.}

\newcommand\minisection[1]{\par\noindent\textbf{#1}\ }

\providecommand{\model}{\mbox{GeoStream}\xspace}

\providecommand{\supp}{\mbox{Supp.~Mat.}\xspace}
\providecommand{\suppmat}{\supp}

\definecolor{Gray}{gray}{0.9}
\definecolor{Celadon}{rgb}{0.67, 0.88, 0.69}
\definecolor{Cream}{rgb}{1.0, 0.99, 0.82}

\definecolor{gold}{HTML}{FAE37F}
\definecolor{silver}{HTML}{D7D7D7}
\definecolor{bronze}{HTML}{EDBA91}

\makeatletter
\newcommand*{\@rowstyle}{}
\newcommand*{\rowstyle}[1]{%
	\gdef\@rowstyle{#1}%
	\@rowstyle\ignorespaces%
}
\newcolumntype{=}{%
	>{\gdef\@rowstyle{}}%
}
\newcolumntype{+}{%
	>{\@rowstyle}%
}
\makeatother

\providecommand{\ie}{i.e.\xspace}

\title{GeoStream: Toward Precise Camera Controlled Streaming Video Generation}

\author{\normalfont
\begin{minipage}{\textwidth}\centering
\bfseries
Yizhou Zhao$^{\lambda}$\thanks{Work partially done during an internship at Snap Inc.} \enspace
Yifan Wang$^{\delta}$ \enspace
Xiaoyuan Wang$^{\lambda}$ \enspace
Yushu Wu$^{\delta}$ \enspace
Hao Zhang$^{\mu}$ \\
Moayed Haji-Ali$^{\rho}$ \enspace
Rameen Abdal$^{\sigma}$ \enspace
Ashkan Mirzaei$^{\sigma}$ \enspace
Yanyu Li$^{\sigma}$ \enspace
Willi Menapace$^{\sigma}$ \\
László A. Jeni$^{\lambda}$ \enspace
Sergey Tulyakov$^{\sigma}$ \enspace
Peter Wonka$^{\kappa,\sigma}$ \enspace
Chaoyang Wang$^{\sigma}$\thanks{Corresponding author.}
\\[0.5em]
\normalfont
$^{\lambda}$ CMU \quad
$^{\delta}$ Northeastern University \quad
$^{\mu}$ UIUC \quad
$^{\rho}$ Rice University \quad
$^{\sigma}$ Snap Inc. \quad
$^{\kappa}$ KAUST
\\[0.5em]
\normalfont
\url{https://a02983278.github.io/geostream.github.io/}
\end{minipage}
}

\begin{document}
\maketitle

\begin{abstract}
	Accurate interactive camera control is essential for video-based world models, but most existing approaches learn camera motion implicitly, leading to inaccurate control under out-of-distribution trajectories.
	Explicit geometric conditioning improves controllability, but existing methods are non-autoregressive and rely on a static 3D cache built from an initial frame, which becomes ineffective once the viewpoint moves beyond the original frustum.
	We propose \model, a framework that enables precise metric-scale camera control in autoregressive streaming video generation.
	Our method maintains a self-refreshing 3D cache that is periodically updated online from the model's own outputs: we estimate depth from the most recently generated frame, unproject to 3D, and reproject into the target view to produce point reprojections as geometric conditioning for subsequent synthesis.
	By the same principle, the conditioning seen during training is also rendered from the student's own generated frames, yielding a fully on-policy distillation that naturally aligns the train and inference conditioning distributions.
	Unlike prior work that uses off-policy condition noising, our approach trains the model against the exact error distribution it encounters at inference, mitigating both standard autoregressive drift and the second-order geometric feedback loop that arises when the cache itself is derived from generated outputs.
	Quantitative and qualitative results show that our approach substantially improves camera controllability.
\end{abstract}

\begin{figure}[t]
	\centering
	\includegraphics[width=0.8\linewidth]{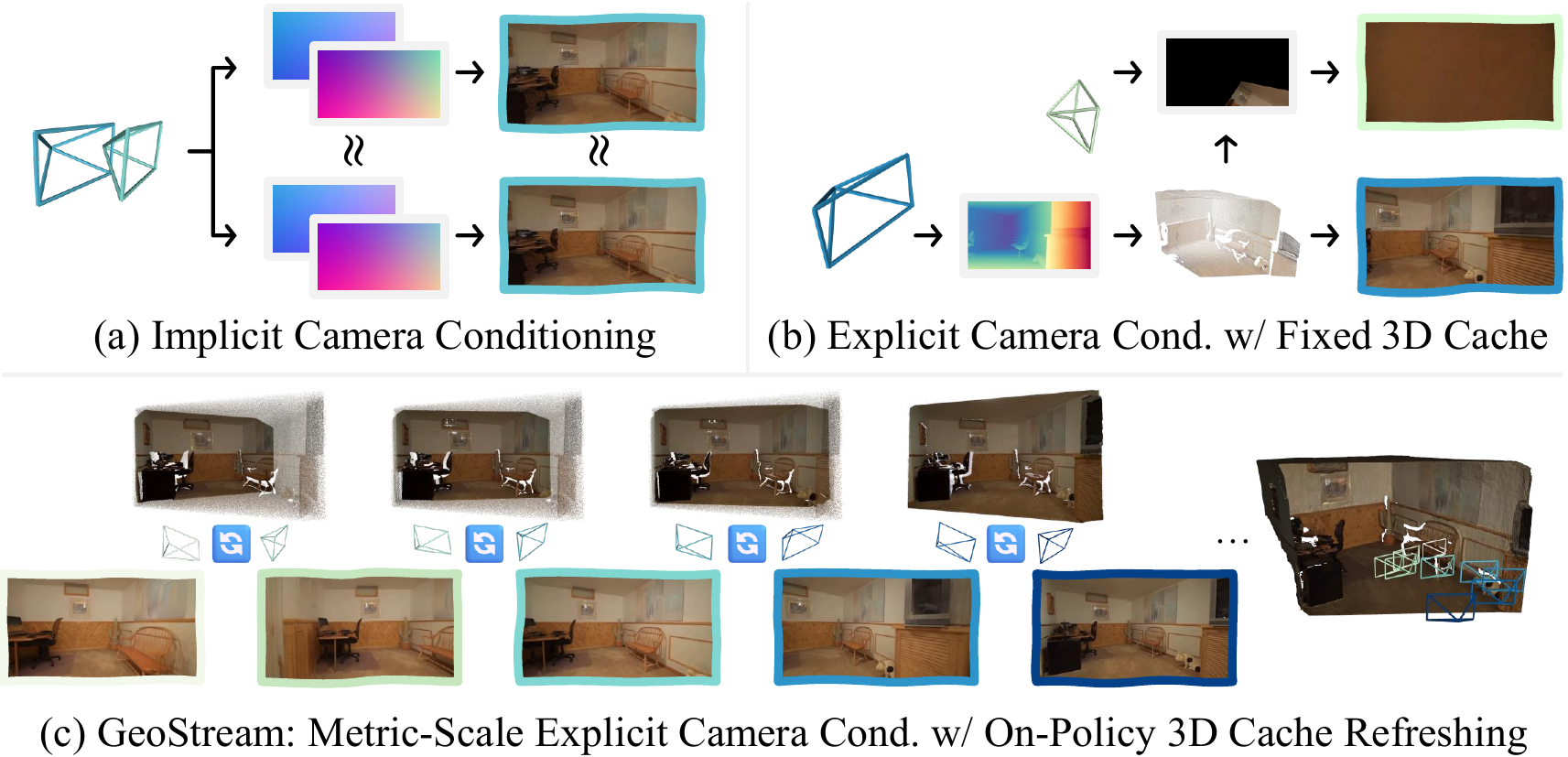}
	\caption{\textbf{Illustration of our motivation.}
		(a)~Implicit camera conditioning lacks geometric constraints, causing the model to regress toward typical or smoothed trajectories rather than following metric-scale motion.
		(b)~Explicit conditioning with a fixed 3D cache becomes ineffective once the viewpoint leaves the original frustum.
		(c)~\model maintains a self-refreshing 3D cache reconstructed from its own generated frames, providing accurate geometric guidance throughout autoregressive generation.}
	\label{fig:intro:teaser}
\end{figure}

\section{Introduction}
Accurate interactive viewpoint control is a core capability of video-based world models.
It enables applications such as immersive scene exploration~\cite{genie,genie3,wang2026one2scenegeometricconsistentexplorable,Paliwal_2025} and robotics simulation~\cite{hafner2023mastering,tu2025drivingworldmodel,robbyantteam2026advancingopensourceworldmodels,ye2026worldactionmodelszeroshot,liu2026worldvlaloopclosedlooplearningvideo,tseng2025scalablepolicyevaluationvideo}.
Most recent video world models operate directly in pixel space~\cite{genie3,he2024cameractrl}.
Camera control is typically implemented by injecting camera pose embeddings~\cite{he2025cameractrl,prope,li2026reroperepurposingroperelative,zhou2024latentreframeenablingcameracontrol,li2025flashworldhighquality3dscene} or adding control layers to the backbone~\cite{he2024cameractrl,wang2024motionctrlunifiedflexiblemotion,wu2025videoworldmodelslongterm}.
While these approaches produce visually plausible motion, they lack explicit geometric grounding.
Camera motion is learned implicitly from data, leading to systematic inaccuracies when the input trajectory deviates from the training distribution.
In particular, increasing the magnitude of input camera translation does not result in proportionally larger motion in the generated video.
As shown in \cref{fig:intro:teaser}a, this scale ambiguity is a fundamental failure mode of implicit control: the model regresses toward typical trajectories rather than following the requested metric-scale motion.
This limitation highlights a missing ingredient: explicit, temporally aligned geometric reasoning during generation.

Another line of work introduces explicit geometric conditioning or a 3D cache.
Methods such as GEN3C~\cite{ren2025gen3c} and TrajectoryCrafter~\cite{yu2025trajectorycrafter} condition generation on projected 3D points or trajectory guidance.
These results show that geometric cues improve viewpoint control and increase data efficiency.
However, these methods are not designed for streaming autoregressive generation.
More importantly, the geometric conditioning is typically derived from a fixed initial frame or an offline reconstruction and remains static during generation.
As illustrated in \cref{fig:intro:teaser}b, without online updates to the 3D cache, these methods become ineffective when viewpoint changes are large~\cite{yu2025trajectorycrafter}, and the precomputed geometry moves out of view.
A parallel line of work pursues long-term 3D scene memory for spatial consistency~\cite{zhao2025spatia,li2025vmem,wu2025videoworldmodelslongterm}.
These methods accumulate and merge geometry across frames.
While effective for global consistency, such accumulation introduces drift and does not target short-term metric-scale camera accuracy under large motions.

These observations suggest a different design principle: accurate camera control requires continually re-estimating geometry from the current generated state, rather than accumulating it over time.

As depicted in \cref{fig:intro:teaser}c, we propose \model, a streaming video generation framework for precise metric-scale camera control, built on a self-refreshing geometric feedback mechanism that remains effective under large camera motions.
Achieving this requires geometry that is both metric and temporally aligned with the generated content, while remaining robust to errors in geometry estimated from generated frames.
We address these challenges with two key components.

First, we introduce a self-refreshing 3D cache for geometric conditioning.
During autoregressive generation, the model periodically reconstructs geometry from its own generated frames.
At each refresh step, we estimate depth from the most recently generated frame, unproject pixels into 3D space, and reproject them into the target camera view.
The resulting point reprojection provides geometrically consistent guidance for subsequent frame synthesis.
Unlike long-term 3D memory, we discard stale geometry at each refresh rather than fusing it, preventing compounding inconsistencies over time.
Because the cache is rebuilt from recently generated frames, it stays aligned with the current content and remains effective under large viewpoint changes.
This design decouples short-term geometric accuracy from long-term content consistency.

Second, we train a causal streaming model that can reliably condition on this self-refreshed cache.
We distill an autoregressive student from a pretrained bidirectional teacher while using the refreshed 3D cache as condition.
This setting is challenging because two forms of exposure bias arise: standard autoregressive drift over long rollouts, and a second-order geometric feedback loop in which errors in generated frames propagate through the depth estimator into the point reprojections and compound over time, an effect invisible to standard self-forcing.
To address this, we adopt a fully on-policy distillation structured in two stages: we first warm up the student off-policy via teacher-forced autoregressive diffusion training to align it with the autoregressive generation distribution, then switch to on-policy DMD distillation in which both the frame stream and the geometric conditioning are rendered from the student's own outputs.
This aligns the training and inference distributions on both channels, jointly mitigating both feedback loops in a single optimization.
In contrast to prior off-policy condition noising~\cite{zhao2025spatia}, which approximates the inference distribution with hand-crafted noise on clean reprojections, our scheme exposes the student to the true self-generated error distribution.

Together, these components enable accurate and stable camera control for streaming autoregressive video generation.
In our evaluation, our method substantially improves camera control accuracy over recent approaches.

In summary, our contributions are:
\begin{itemize}
	\item We propose \model, a streaming autoregressive video generation framework with explicit, online-refreshed 3D geometric conditioning, enabling precise viewpoint manipulation under large camera motions.
	\item We introduce a periodic 3D cache refresh mechanism that maintains temporally aligned geometric conditioning under large camera motions, without requiring long-term scene memory or offline reconstruction.
	\item We develop a fully on-policy training strategy structured as off-policy warmup followed by on-policy DMD distillation, that mirrors the inference-time self-refresh of the 3D cache, eliminating geometry-induced exposure bias by exposing the student to the true self-generated conditioning distribution, differentiating from off-policy condition-noising approaches.
\end{itemize}

\section{Related Work}

\begin{figure*}[t]
	\centering
	\includegraphics[width=\linewidth]{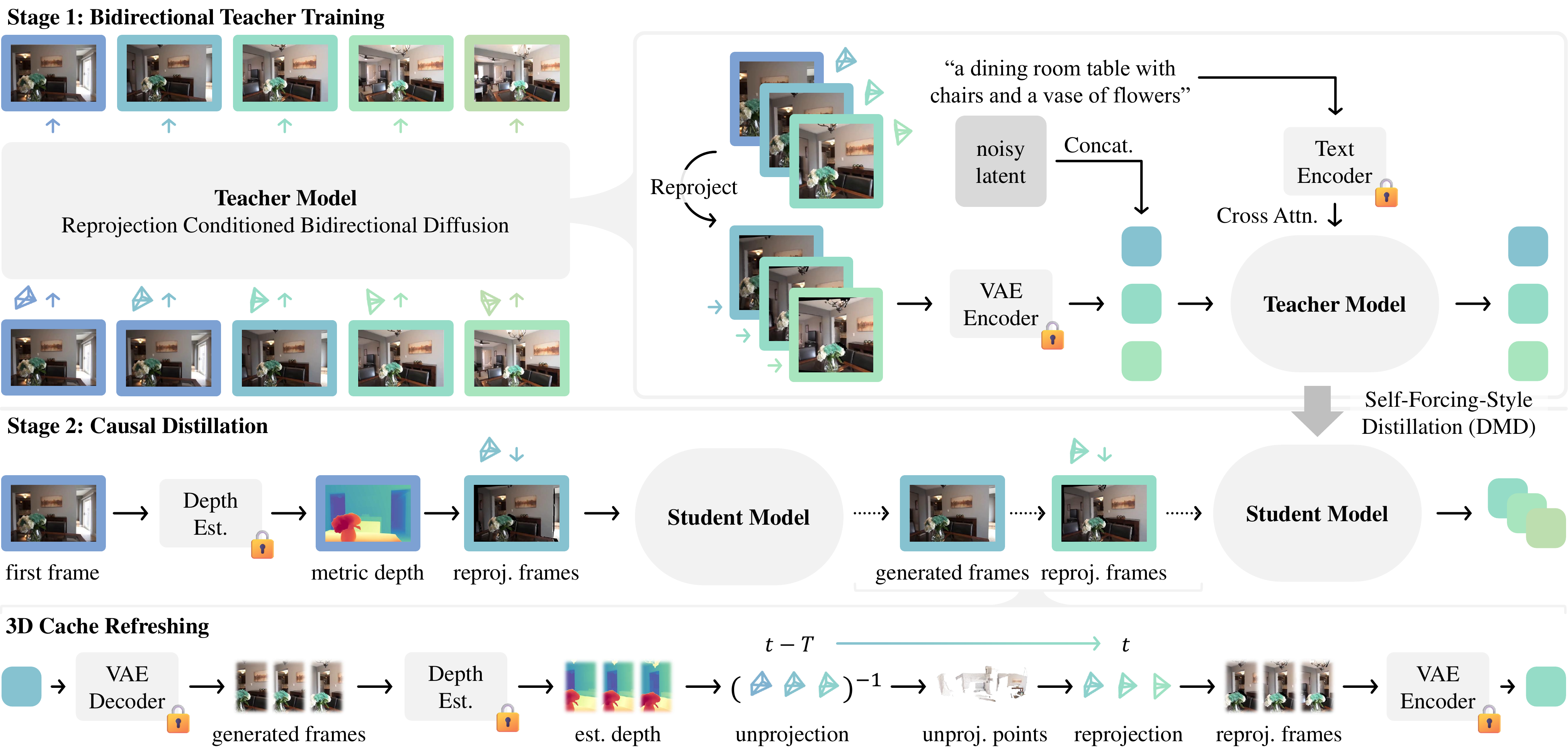}
	\caption{
		\textbf{Overview of \model pipeline.}
		The pipeline consists of two stages.
		Stage 1: Bidirectional Teacher Training.
		A teacher model is trained with ground-truth~(GT) frames and explicit geometric conditioning via point reprojection.
		Stage 2: Causal Distillation.
		A causal student is first warmed up off-policy via autoregressive diffusion training~\cite{zhu2026causalforcing} and then distilled on-policy via Distribution Matching Distillation (DMD)~\cite{dmd} using the bidirectional teacher as the score model, with both frames and point reprojection conditioning rendered from the student's own outputs to mitigate geometric exposure bias.
		To handle large camera motions during inference, the student utilizes a \emph{Self-Refreshed 3D Cache} that periodically updates the geometry using depth estimated from its own generated frames.
	}
	\label{fig:method:pipeline}
\end{figure*}

\subsection{Camera Controllable Video Generation}

Camera controllable video generators are either implicit, injecting the camera signal as per-frame pose embeddings~\cite{he2025cameractrl,he2024cameractrl,Yu_2025_ICCV,prope,li2026reroperepurposingroperelative,zhou2024latentreframeenablingcameracontrol,li2025flashworldhighquality3dscene} or grafting auxiliary control layers onto a bidirectional backbone~\cite{wang2024motionctrlunifiedflexiblemotion,wu2025videoworldmodelslongterm}, or explicit, conditioning on a colored 3D point cloud reprojected into each target view~\cite{ren2025gen3c,yu2024viewcrafter,yu2025trajectorycrafter,cao2025uni3c,lee20253d}.
Implicit methods learn camera motion as a statistical correlation, yielding plausible motion but leaving two systematic failure modes: high-frequency motions are smoothed out, and translation scale is ambiguous because the training objective is largely scale-invariant once scenes are normalized.
Explicit methods resolve both, but the cache is almost always built once from an initial frame.
As soon as the viewpoint leaves the original frustum, the geometric signal degenerates into a noisy, partially-observed cue.
Some variants maintain persistent spatial memory~\cite{zhao2025spatia} or extend toward 4D scene modeling~\cite{yang2026neoverse,xu2025virtuallybeing,huang2025voyager,song2025worldforgeunlockingemergent3d4d,zheng2026versecrafterdynamicrealisticvideo,sun2025unigeo,yao2025uni4d,xie2026lavr,wu2025geometry,jiang2025geo4d,chou2025flashdepth,hu2025ex4dextremeviewpoint4d}, targeting long-term scene consistency, which is a complementary goal to ours.
Our work inherits the explicit-cache philosophy but makes it genuinely streaming: rebuilt online at every chunk boundary without offline reconstruction, discarding stale geometry at each refresh rather than fusing it to avoid compounding inconsistencies from merging feed-forward depth estimates over long horizons.
Unlike Spatia~\cite{zhao2025spatia}, which updates conditioning off-policy and never exposes the model to the fully self-generated error distribution, our approach is fully on-policy, directly closing the second-order feedback loop.

\subsection{Streaming Video Generation}

A second thread reformulates video diffusion as a causal, chunk-by-chunk rollout~\cite{chen2024diffusion,songhistory,gu2025diffusionshader3dawarevideo,mittal2025uniphy,ai2025magi1autoregressivevideogeneration} for real-time generation.
Converting a bidirectional model to causal introduces a train-test mismatch: at inference every context token comes from the model itself, whereas at training it is oracle ground truth.
Distillation recipes such as asymmetric distribution matching~\cite{yin2025slow} and self-forcing~\cite{huang2025self} close this gap by rolling the student out on its own trajectory.
Long-horizon variants compress or re-anchor the KV cache~\cite{yang2025longlive,cui2025self,liu2025rolling}, efficient implementations target mobile deployment~\cite{zhao2026s2ditsandwichdiffusiontransformer,cui2026lol,dai2025fantasyworld}, and a parallel direction integrates multi-modal control signals into the streaming pipeline~\cite{shin2026motionstream,robbyantteam2026advancingopensourceworldmodels,liu2025worldweavergeneratinglonghorizonvideo,li2025vmem,geng2025motion,huang2025memory}.
While these advances address textual and appearance exposure bias, none model the geometric exposure bias that arises when an explicit 3D cache is coupled with autoregressive generation: errors in generated frames propagate through the depth estimator into subsequent conditioning, forming a compounding feedback loop invisible to standard self-forcing.

\subsection{Geometrical Estimation}

Because our cache is rebuilt online from a single generated frame, we require a monocular geometry estimator that is both fast and metric.
Recent monocular depth predictors substantially improve robustness and metric fidelity~\cite{depth_anything_v2,hu2024metric3d,piccinelli2025unidepthv2,wang2025moge,zhang2025tapip3d,Byung-Ki_2025_ICCV,dong2025one}, and video depth models~\cite{video_depth_anything} extend these to temporally consistent prediction.
Feed-forward reconstructors such as VGGT~\cite{wang2025vggt}, $\pi^3$~\cite{wang2025pi3permutationequivariantvisualgeometry}, and MapAnything~\cite{keetha2025mapanything} jointly infer cameras and geometry from one or multiple views~\cite{zhang2024worldconsistentvideodiffusionexplicit,deng2025vggtlongchunkitloop,chen2025reconstruct}.
Streaming reconstructors~\cite{wang2025continuous,streamVGGT,chen2025ttt3r,yuan2026infinitevggt,feng2025st4rtrack} and streaming splatters~\cite{wu2025streamsplat,lin2025movies} maintain a persistent scene state.
Among these, we adopt MoGe-2~\cite{wang2025moge} as our geometry backbone because it provides better geometry quality and more accurate scale consistency, thereby facilitating precise metric-scale camera control in our framework.

\begin{figure*}[t]
	\centering
	\includegraphics[width=\linewidth]{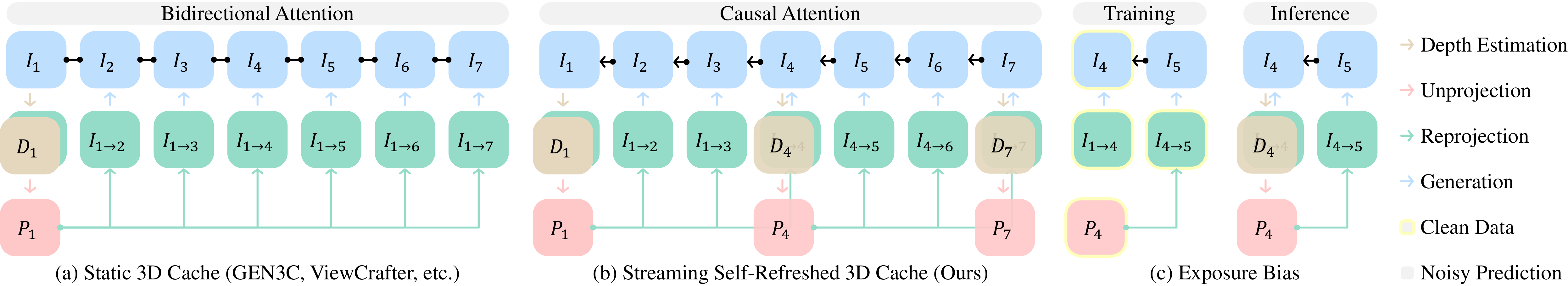}
	\caption{\textbf{Static vs. self-refreshed 3D cache.}
		(a) Static 3D Cache uses a fixed initial cache $\mathcal{P}_1$, which loses overlap as the camera moves.
		(b) Streaming Self-Refreshed 3D Cache periodically rebuilds geometry from recent frames, maintaining alignment over long horizons.
		(c) Exposure Bias: at training, the context $I_4$ and cache $\mathcal{P}_4$ are clean ground truth.
		At inference, $\mathcal{P}_4$ is unprojected from depth re-estimated on the model's own noisy $I_4$, so the conditioning $I_{4\rightarrow 5}$ deviates from training.
		This second-order feedback loop on the cache compounds over the rollout, is invisible to standard self-forcing, and is not resolved by off-policy condition noising on clean reprojections~\cite{zhao2025spatia}.}
	\label{fig:method:illustration}
\end{figure*}

\section{Method}

Our goal is to train an autoregressive video model that generates frames interactively under user-specified viewpoint control.
The model should follow the input camera trajectory accurately while maintaining temporal consistency.
To achieve precise control, we condition generation on explicit geometric constraints at inference time, with an overview shown in \cref{fig:method:pipeline}.
These constraints are represented as a 3D cache that is periodically refreshed to support large camera motions (\cref{sec:method:inference}).
To make this inference mechanism work in an autoregressive setting, we propose a training recipe that mitigates exposure bias introduced by the 3D cache and reduces temporal drift (\cref{sec:method:training}).
The two components share a single principle: the geometric conditioning is always computed from the student's own generated frames, \emph{self-refreshed} at inference and \emph{self-rolled-out} during distillation, so that the train and inference distributions of the conditioning signal are aligned by nature.

\subsection{Self-Refreshed 3D Cache}
\label{sec:method:inference}

\minisection{Unprojected 3D points as 3D cache.}
Following prior work~\cite{ren2025gen3c,yu2024viewcrafter}, we represent the 3D cache as a dense colored point cloud obtained by unprojecting a depth map from a previous frame.
Given a frame $I_t$ at time $t$, we estimate its depth map $D_t$ and unproject it to form a point cloud $\mathcal{P}_t$, which serves as the 3D cache.
To generate a subsequent frame $I_s$ ($s>t$), we project $\mathcal{P}_t$ into the target view using the input camera projection matrix $M_s$ and render a partial image $I_{t\rightarrow s}$.
The video model then uses $I_{t\rightarrow s}$ as a geometric conditioning signal to synthesize the full frame $I_s$.

\minisection{Sliding-window cache refreshing.}
Illustrated in \cref{fig:method:illustration}(a), prior approaches~\cite{ren2025gen3c,yu2024viewcrafter} construct a static cache from only the first frame.
As the viewpoint changes, the overlap between the cached geometry and the current viewing frustum decreases, reducing the effectiveness of the conditioning.

We instead refresh the 3D cache during autoregressive generation, as in \cref{fig:method:illustration}(b).
We split the sequence into non-overlapping temporal windows of size $c$, where $c$ denotes the autoregressive chunk size, i.e., the number of frames generated per autoregressive step, and refresh the cache at the end of each window.
Concretely, at frame $t{=}1+kc$ for $k=1,2,\ldots$, we replace the cache with the unprojected point cloud $\mathcal{P}_{1+kc}$ estimated from that frame.
Frames in the next window are then generated conditioned on this refreshed cache.
We find this simple strategy is sufficient to support large viewpoint changes while preserving precise camera controllability.

Unlike prior work that performs online point cloud fusion~\cite{zhao2025spatia}, we do not merge caches across windows.
We always discard the previous cache upon refresh to avoid inconsistencies such as layering artifacts, which can arise when merging point clouds produced by feed-forward reconstruction methods~\cite{wang2025vggt}.
Instead, we rely on the video model's implicit spatial memory to maintain long-horizon consistency, while using the explicit 3D cache primarily to improve camera control accuracy.

We set the refresh window size to match the number of frames generated per autoregressive chunk.
This refreshes the cache as frequently as possible without introducing additional overhead.

\minisection{3D cache conditioning.}
We adapt a pretrained video diffusion model to accept point reprojection images rendered from the 3D cache as a geometric conditioning signal.
To inject this conditioning into the video DiT~\cite{dit}, we use channel concatenation for efficiency rather than sequence concatenation.
We encode the point reprojection video using the causal encoder of a pretrained video VAE, producing low-resolution latent maps.
These latents are patchified and linearly projected into hidden states $z_\text{3D}$ with the same dimensionality as the noised latents $z_\text{video}$.
We fuse the two streams via element-wise addition, \ie $z_\text{video}+z_\text{3D}$, which is equivalent to channel concatenation followed by a linear projection into the original hidden dimension.
We keep the rest of the transformer architecture unchanged and train it to predict the velocity for denoising the corrupted video latents $z_\text{video}$.
During training, we update only the projection layers for $z_\text{3D}$ and the self-attention layers in each transformer block.

\begin{figure*}[t]
	\centering
	\includegraphics[width=\linewidth]{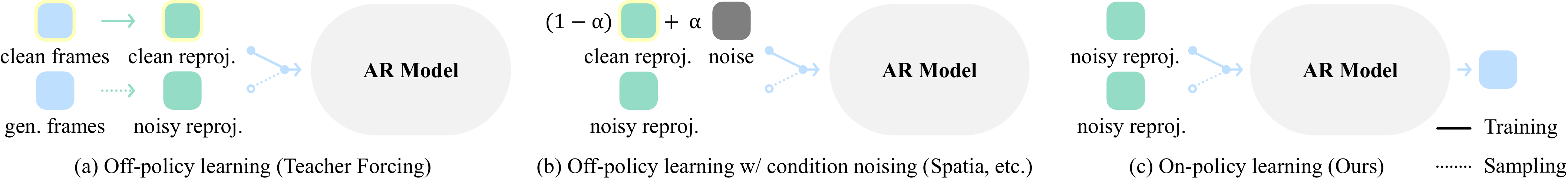}
	\caption{\textbf{Training schemes for autoregressive generation.}
		(a) Off-policy learning, e.g., Teacher Forcing: The model is trained on clean ground-truth reprojections but sampled using noisy generated ones, leading to exposure bias.
		(b) Off-policy learning with condition noising, e.g., Spatia~\cite{zhao2025spatia}: To bridge the train-test gap, random noise is injected into the clean condition latent during training as a data augmentation.
		(c) On-policy learning: After off-policy warmup, our model is distilled on-policy via DMD using reprojections rendered from its own generated frames, aligning the training and sampling distributions on both the frame and geometric conditioning channels, directly closing the second-order feedback loop.}
	\label{fig:method:solution}
\end{figure*}

\subsection{On-Policy Causal Distillation}
\label{sec:method:training}

Distilling a causal video diffusion model under periodically refreshed 3D cache introduces two major sources of exposure bias, as illustrated in \cref{fig:method:illustration}(c).
First, there is the standard autoregressive exposure bias.
During training, previous frames are ground-truth frames.
During inference, all previous frames are generated and may contain accumulated errors.
This mismatch leads to quality degradation over long rollouts.
Second, the geometric conditioning itself is self-generated at inference time.
The point reprojection is computed from the depth estimated on previously generated frames.
Errors in texture or geometry, therefore, propagate into the point reprojections.
Imperfect reprojections further degrade future generations, creating a compounding feedback loop.
To address these issues, we introduce the following strategies.

\minisection{Off-policy initialization with teacher forcing.}
We first train a bidirectional video diffusion teacher conditioned on point reprojection frames computed from ground-truth depth and ground-truth frames.
We then initialize the causal student off-policy via teacher-forced autoregressive diffusion training~\cite{zhu2026causalforcing} as in \cref{fig:method:solution}(a): the model is fine-tuned in a causal setting where both the context frames and the point reprojection conditioning are taken from ground truth, and the student is trained to denoise the current chunk given this clean context.
This warm-up aligns the student with the autoregressive generation distribution and provides a stable initialization for the subsequent on-policy stage, but by design, it leaves both sources of exposure bias unaddressed: the student never observes its own previously generated frames, nor the artifact-laden reprojections rendered from them.

\minisection{On-policy distillation with self-rollout condition.}
We then distill the causal student on-policy via DMD~\cite{dmd,huang2025self} so that both forms of exposure bias are closed jointly as illustrated in \cref{fig:method:solution}(c).
The key observation is that training-time \emph{self-rollout} of the conditioning is the natural counterpart of the inference-time \emph{self-refresh} introduced in \cref{sec:method:inference}: in both stages, the point reprojection is rendered from the student's own generated frames, so by adopting the same mechanism during distillation, we naturally close the train-inference gap on the conditioning channel.
Concretely, at each training step, the student rolls out a streaming video sequence from its own outputs, and we re-render the point reprojection conditioning from depth estimated on these self-generated frames rather than on the ground truth.
The student is supervised by a DMD loss using the bidirectional teacher as the score model.
By coupling self-rollout on the frame stream with self-rollout on the geometric conditioning, the student is supervised against the same joint distribution of context frames and reprojection artifacts it encounters at inference, jointly mitigating the standard autoregressive exposure bias and the second-order geometric feedback loop introduced by the self-refreshed 3D cache in a single optimization.
In contrast to off-policy condition noising~\cite{zhao2025spatia}, which approximates this distribution with hand-crafted noise on clean reprojections as illustrated in \cref{fig:method:solution}(b), our scheme exposes the student to the true error distribution and lets the DMD loss absorb both feedback loops together.

\section{Experiments}

\subsection{Experimental Settings}
\label{sec:experiment:settings}

\begin{table}[t]
	\footnotesize
	\centering
	\caption{\textbf{Quantitative comparison with state-of-the-art methods.}
		We evaluate our method against existing approaches using video quality metrics (FVD, PSNR, SSIM, LPIPS) and camera trajectory accuracy metrics (ATE, RTE, RRE).
		The lower block (\textit{w/ MoGe-2}) re-evaluates baselines whose monocular depth backbone is swappable, replacing it with the same MoGe-2~\cite{wang2025moge} estimator used by our method, in order to factor out gains attributable to the depth backbone alone.
		$\uparrow$ ($\downarrow$) indicates that higher (lower) values are better.
		The best results are highlighted in bold.}
	\resizebox{\linewidth}{!}{
		\begin{tabular}{lccccccc}
			\toprule
			Method                        & FVD$\downarrow$ & PSNR$\uparrow$ & SSIM$\uparrow$ & LPIPS$\downarrow$ & ATE$\downarrow$ & RTE$\downarrow$ & RRE$\downarrow$ \\
			\midrule
			MotionCtrl                    & 853.8           & 9.17           & 0.432          & 0.809             & 9.051           & 2.373           & 21.757          \\
			SEVA                          & 433.1           & 12.14          & 0.499          & 0.618             & 1.220           & 0.071           & 0.299           \\
			CameraCtrl II                 & 207.9           & 14.51          & 0.601          & 0.472             & 3.549           & 0.189           & 0.275           \\
			FlexWorld                     & 747.6           & 11.10          & 0.484          & 0.702             & 9.620           & 0.458           & 3.067           \\
			ViewCrafter                   & 1268.9          & 7.80           & 0.193          & 0.856             & 3.220           & 0.551           & 7.707           \\
			VMem                          & 555.0           & 10.65          & 0.446          & 0.759             & 7.10            & 0.949           & 12.584          \\
			GEN3C                         & 218.5           & 13.71          & 0.545          & 0.514             & 6.020           & 0.176           & 0.264           \\
			Spatia                        & 190.4           & 14.48          & 0.533          & 0.483             & 1.140           & 0.060           & 0.580           \\
			\midrule
			FlexWorld w/ MoGe-2           & 133.9           & 16.16          & 0.614          & 0.420             & 0.911           & 0.046           & 0.247           \\
			ViewCrafter w/ MoGe-2         & 249.9           & 14.71          & 0.622          & 0.463             & \textbf{0.539}  & 0.064           & 0.241           \\
			VMem w/ MoGe-2                & 417.4           & 12.62          & 0.562          & 0.609             & 1.537           & 0.254           & 1.378           \\
			GEN3C w/ MoGe-2               & 385.8           & 12.32          & 0.559          & 0.651             & 3.977           & 0.166           & 1.373           \\
			Spatia w/ MoGe-2              & 163.9           & 15.95          & 0.629          & 0.433             & 0.946           & 0.051           & 0.295           \\
			\rowcolor{Gray} \model (Ours) & \textbf{98.1}   & \textbf{17.59} & \textbf{0.651} & \textbf{0.336}    & 0.816           & \textbf{0.032}  & \textbf{0.162}  \\
			\bottomrule
		\end{tabular}
	}
	\label{tab:experiment:comparison}
\end{table}

\begin{figure*}[t]
	\centering
	\includegraphics[width=\linewidth]{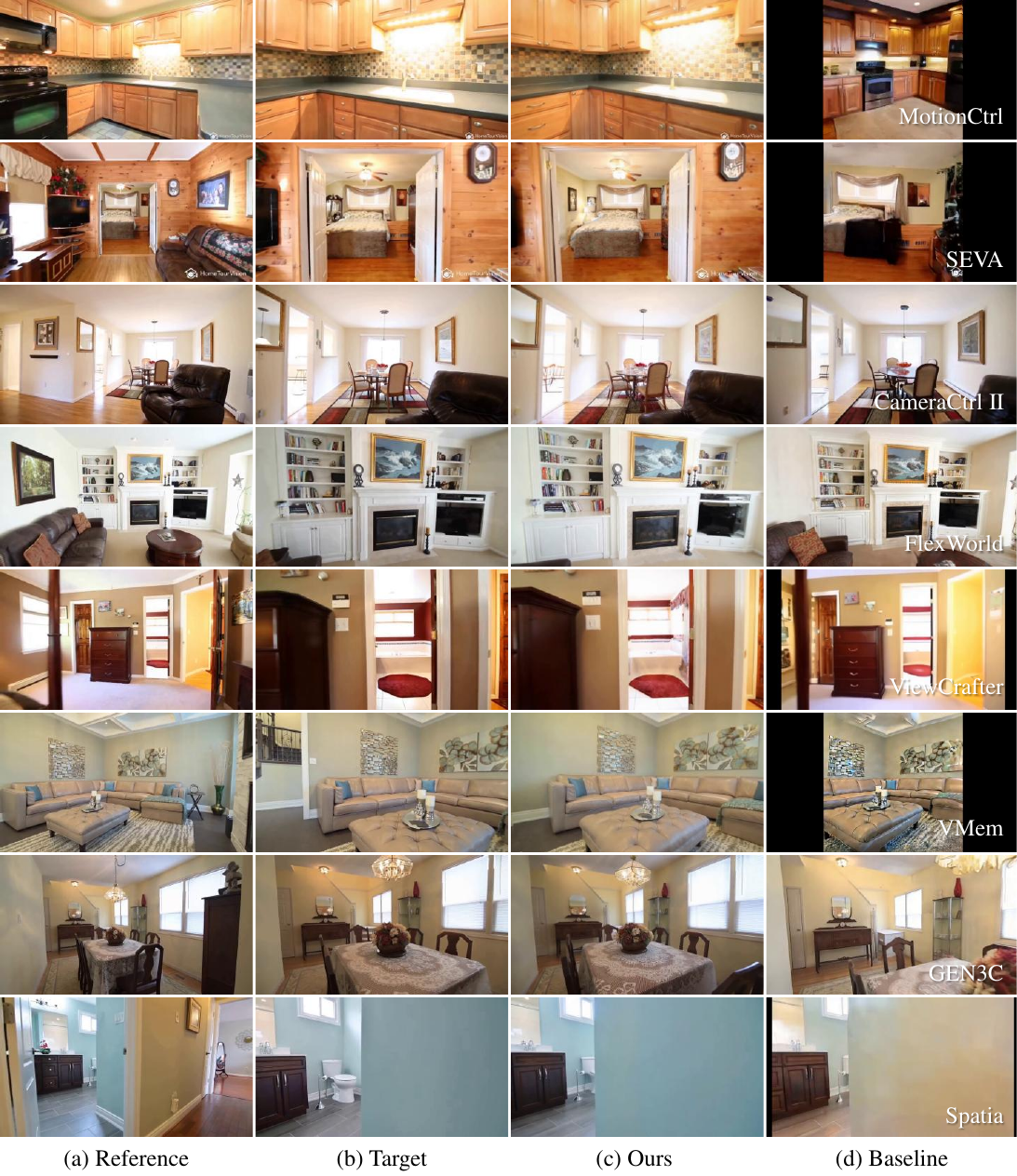}
	\caption{\textbf{Qualitative comparison of camera-controlled video generation.}
		Compared with state-of-the-art baselines, our \model generates videos with higher visual fidelity and stricter adherence to the input camera trajectories.}
	\label{fig:experiment:comparison}
\end{figure*}

\minisection{Implementation details.}
We build both the bidirectional teacher and the causal student on top of Wan2.2-5B~\cite{wan2025} with spatial resolution of $1280 \times 704$ and 81 frames.
Our training proceeds in three stages.
First, we fine-tune the bidirectional teacher using AdamW with a learning rate of $1\times10^{-5}$ for 8K iterations.
Second, we warm up the causal student via autoregressive diffusion training~\cite{zhu2026causalforcing} for 5K iterations with a learning rate of $2\times10^{-6}$.
Finally, we perform on-policy DMD distillation for 1K iterations.
In this stage, we adopt a $1\times10^{-5}$ learning rate for the student model, and a $2\times10^{-6}$ learning rate for the fake score model.
Following Self-Forcing~\cite{huang2025self} and MotionStream~\cite{shin2026motionstream}, the sparse causal attention pattern of the student is parameterized by three quantities.
The \emph{chunk size} $c$ denotes the number of frames generated per autoregressive step.
The \emph{sink size} $s$ is the number of leading chunks always kept in the KV cache.
The \emph{local context window size} $w$ is the number of most recent past chunks attended to.
We set $c=4$, $s=1$, $w=3$ throughout all comparisons, following the ablation analysis in \cref{fig:experiment:chunk_sink_window} of \suppmat.

\minisection{Datasets.}
We train and evaluate our model on RealEstate10K~\cite{realestate10k}, a dataset consisting of real-estate walkthrough videos with per-frame poses and trajectories estimated via SLAM and bundle adjustment; our quantitative validation accordingly focuses on static scenes.
While the raw clips vary from dozens to hundreds of frames in length at a native $1024 \times 576$ resolution, we trim them to 81 frames and resize them to the specific input requirements of each compared baseline, spanning 256P to 720P, to ensure a fair quantitative comparison.
This fixed frame count accommodates baselines involving intensive geometric reconstruction for our primary evaluations, whereas our ablation studies in \suppmat further demonstrate the stability of our autoregressive approach under longer roll-outs.

\minisection{Baselines.}
We conduct a close comparison with recent camera-controlled video generation models.
Methods without explicit geometric constraints include MotionCtrl~\cite{wang2024motionctrlunifiedflexiblemotion}, SEVA~\cite{zhou2025seva}, and CameraCtrl II~\cite{he2025cameractrl} in \cite{wan2025}.
Methods that leverage explicit geometric constraints include FlexWorld~\cite{li2025flexworld}, ViewCrafter~\cite{yu2024viewcrafter}, VMem~\cite{li2025vmem}, GEN3C~\cite{ren2025gen3c}, and Spatia~\cite{zhao2025spatia}.

\minisection{Evaluation metrics.}
We evaluate our method across two key dimensions: visual synthesis quality and camera control ability.
We adopt Fréchet Video Distance (FVD) to characterize the overall temporal coherence and distribution similarity of the generated sequences, where features are extracted using an I3D backbone at a resolution of $224 \times 224$.
For frame-level fidelity, we report PSNR, SSIM, and LPIPS~\cite{zhang2018perceptual} to measure structural and perceptual realism against the ground truth.
To quantify camera accuracy, we reconstruct metric-scale trajectories from the generated videos using MapAnything~\cite{keetha2025mapanything} for comparison against the reference control signals.
They are further analyzed via Absolute Trajectory Error (ATE) for global consistency, complemented by Relative Translation Error (RTE) and Relative Rotation Error (RRE) for local drift assessment.

\subsection{Comparison Results}

\minisection{Quantitative and qualitative comparison.}
As shown in the upper block of \cref{tab:experiment:comparison}, when each baseline is evaluated with its original geometry estimator, \model is significantly more accurate than all competing methods on the camera-pose metrics ATE, RTE, and RRE, and also leads on visual fidelity across FVD, PSNR, SSIM, and LPIPS.
We emphasize that \model is the \textbf{only autoregressive} entry in this comparison.
All other methods are bidirectional, a strictly easier setting for both video synthesis and camera control, since the model can attend to future frames and amortize geometry over the entire trajectory.
Qualitative comparisons in \cref{fig:experiment:comparison} corroborate the numbers: our outputs closely match the ground truth, whereas VMem~\cite{li2025vmem} produces distorted structure, ViewCrafter~\cite{yu2024viewcrafter} blurry ceilings, MotionCtrl~\cite{wang2024motionctrlunifiedflexiblemotion} black boundaries, and CameraCtrl II~\cite{he2025cameractrl} yields imprecise viewpoints.

\begin{table}[t]
	\footnotesize
	\centering
	\caption{\textbf{Ablation study on geometry estimation methods.}
		We investigate the influence of various geometry priors within our framework while keeping the backbone and resolution constant.
		The geometry ablation is conducted under the off-policy training setup, whereas \cref{tab:experiment:policy} fixes the geometry to MoGe-2 and varies the training scheme.}
	\label{tab:experiment:geometry}
	\resizebox{0.85\linewidth}{!}{
		\begin{tabular}{lccccccc}
			\toprule
			Variants                  & FVD$\downarrow$ & PSNR$\uparrow$ & SSIM$\uparrow$ & LPIPS$\downarrow$ & ATE$\downarrow$ & RTE$\downarrow$ & RRE$\downarrow$ \\
			\midrule
			w/ InfiniteVGGT           & 299.6           & 11.87          & 0.534          & 0.661             & 5.275           & 0.589           & 3.260           \\
			w/ MapAnything            & 158.3           & 15.72          & 0.597          & 0.436             & \textbf{0.917}  & \textbf{0.062}  & \textbf{0.270}  \\
			\rowcolor{Gray} w/ MoGe-2 & \textbf{126.3}  & \textbf{16.90} & \textbf{0.631} & \textbf{0.390}    & 1.060           & 0.116           & 0.356           \\
			\bottomrule
		\end{tabular}
	}
\end{table}

\begin{table}[t]
	\footnotesize
	\centering
	\caption{\textbf{Ablation study on training schemes.} We compare the performance of different learning strategies: off-policy, off-policy with condition noising, and our on-policy learning.}
	\label{tab:experiment:policy}
	\resizebox{\linewidth}{!}{
		\begin{tabular}{lccccccc}
			\toprule
			Variants                        & FVD$\downarrow$ & PSNR$\uparrow$ & SSIM$\uparrow$ & LPIPS$\downarrow$ & ATE$\downarrow$ & RTE$\downarrow$ & RRE$\downarrow$ \\
			\midrule
			Off-policy                      & 126.3           & 16.90          & 0.631          & 0.390             & 1.060           & 0.116           & 0.356           \\
			Off-policy w/ condition noising & 123.2           & 16.35          & 0.646          & 0.379             & 1.090           & 0.076           & 0.172           \\
			\rowcolor{Gray} On-policy       & \textbf{98.1}   & \textbf{17.59} & \textbf{0.651} & \textbf{0.336}    & \textbf{0.816}  & \textbf{0.032}  & \textbf{0.162}  \\
			\bottomrule
		\end{tabular}
	}
\end{table}

\minisection{Controlling for the depth backbone.}
A natural concern is whether our gains simply reflect a stronger depth estimator.
To isolate this factor, we re-evaluate FlexWorld, ViewCrafter, VMem, GEN3C, and Spatia by replacing their geometry backbone with the same MoGe-2~\cite{wang2025moge} estimator used by \model, shown in the lower block of \cref{tab:experiment:comparison}.
Swapping in MoGe-2 cuts FlexWorld's and ViewCrafter's FVD by roughly $80\%$, but is not uniformly beneficial: GEN3C's FVD even degrades, since each baseline was trained jointly with its original geometry conditioning and does not necessarily transfer cleanly to a swapped estimator at test time.
Even so, the depth backbone is clearly a non-trivial factor, and \model still leads the strongest \textit{w/ MoGe-2} baseline on FVD, all frame-level fidelity metrics, and the local trajectory metrics RTE and RRE.
The only exception is global ATE, on which ViewCrafter w/ MoGe-2 ($0.539$) benefits from generating the full trajectory at once non-causally, while \model operates strictly causally on its own previously generated frames.

\subsection{Ablation Study}

\minisection{Geometrical estimation.}
We evaluate the impact of different geometric priors on RealEstate10K as shown in \cref{tab:experiment:geometry}.
While MapAnything provides competitive pose accuracy, MoGe-2 emerges as the superior backbone across nearly all metrics.
The performance gain is primarily attributed to MoGe-2's explicit decoupling of relative geometry from global scale prediction, which effectively resolves the inherent focal-distance ambiguity in monocular sequences.
By minimizing geometric distortion during the reprojection process, MoGe-2 ensures higher structural integrity and temporal coherence in the generated videos.

\minisection{Training scheme.}
We compare three distillation strategies under the same self-refreshed 3D cache in \cref{tab:experiment:policy}.
\emph{Off-policy} learns from clean reprojections built from ground-truth depth and frames.
\emph{Off-policy w/ condition noising}~\cite{zhao2025spatia} adds random noise to those clean reprojections as a data augmentation.
Our \emph{on-policy} scheme instead renders the point projection on the fly from the student's own generated frames.
Relative to plain off-policy, condition noising yields only a marginal gain, with FVD dropping by $2.5\%$ and ATE roughly unchanged, whereas going fully on-policy improves all metrics substantially, reducing FVD by $22.3\%$, ATE by $23.0\%$, and RTE by $72.4\%$, with the largest gains on local trajectory metrics where geometric exposure bias compounds the fastest.
This confirms that closing the second-order feedback loop, not just smoothing the conditioning manifold, is the critical factor.

\begin{figure}[t]
	\centering
	\includegraphics[width=0.9\linewidth]{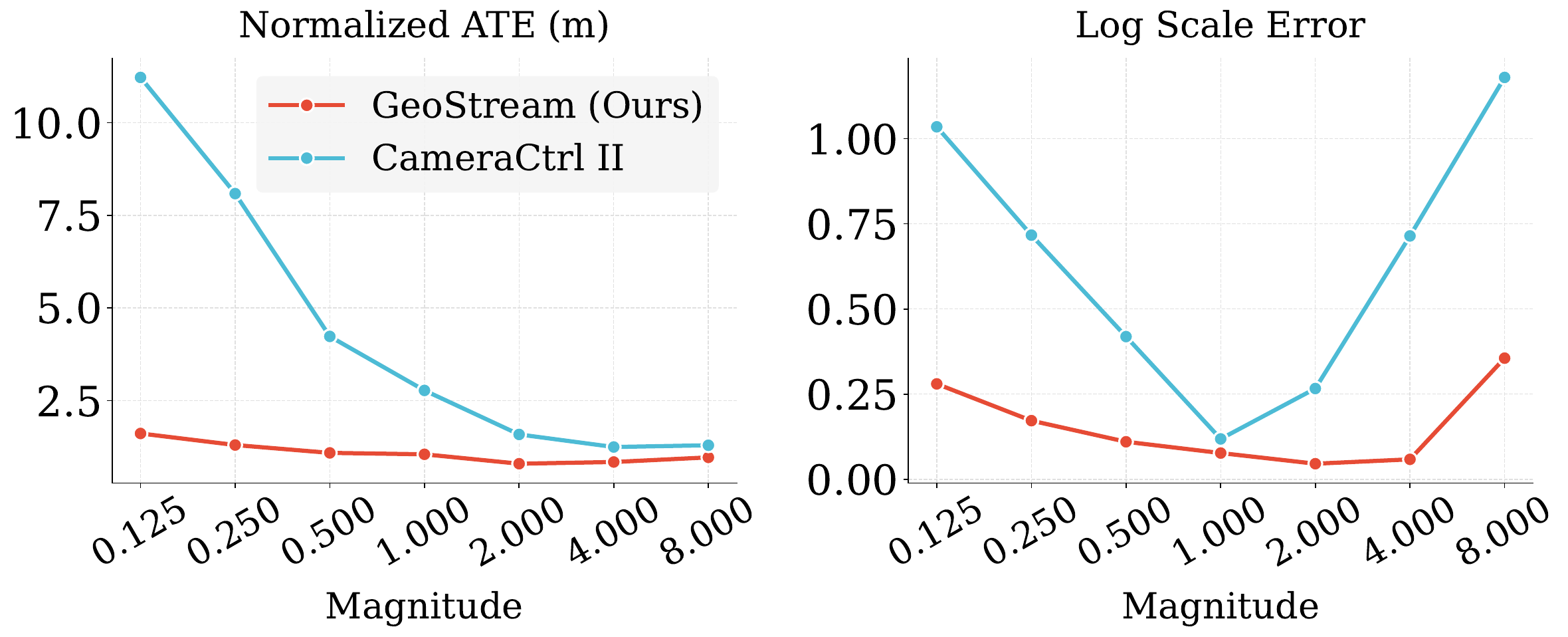}
	\caption{\textbf{Impact of motion magnitude on control accuracy.}
		We scale camera translation by factor $m$.
		\model consistently maintains lower error than CameraCtrl~II.
		Normalized ATE and Log Scale Error highlight our superior fine-grained control accuracy and scale preservation ability, especially at small magnitudes ($m < 1$).}
	\label{fig:experiment:magnitude}
\end{figure}

\begin{figure}[t]
	\centering
	\includegraphics[width=0.95\linewidth]{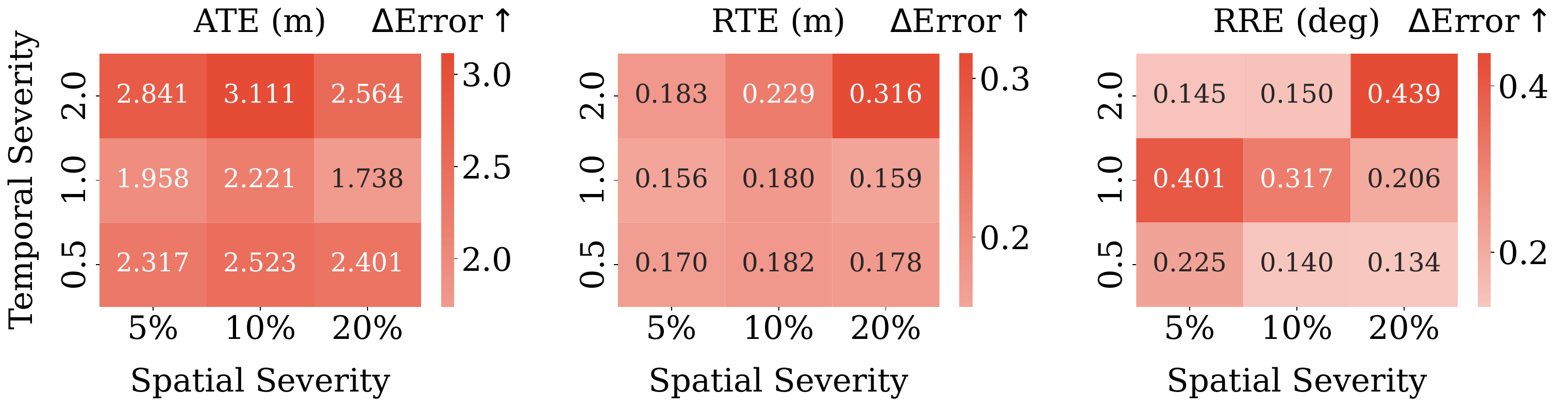}
	\caption{\textbf{Robustness to camera trajectory perturbations.}
		We evaluate control accuracy under sinusoidal noise with varying spatial magnitude and temporal frequency.
		Heatmaps visualize $\Delta\text{Error} = \text{CameraCtrl II} - \text{Ours}$, where red indicates \model superiority.
		\model consistently reduces all metrics across various noise levels.}
	\label{fig:experiment:severity}
\end{figure}

\subsection{Quantitative Analysis}
\label{sec:experiment:quantitative}

\minisection{Sensitivity to camera translation magnitude.}
A clean way to expose whether a generator has truly learned metric camera motion, or only its direction, is to rescale the input trajectory and ask the same question at different magnitudes.
We sweep magnitude factor $m\!\in\!\{0.125, 0.25, 0.5, 1, 2, 4, 8\}$ and plot three views in \cref{fig:experiment:magnitude}.
Once we normalize by $m$, \model's curve is nearly flat, whereas CameraCtrl II's explodes at $m\!<\!1$, showing the implicit controller cannot resolve small motions that lie below the scale granularity of its training distribution.
The Log-Scale-Error plot on the right exposes the underlying mechanism.
CameraCtrl II systematically overshoots small requests and undershoots large ones, yielding a characteristic ``regression-to-the-mean'' curve that is the hallmark of a scale-ambiguous prior, where the network has effectively learned a distribution over plausible trajectories and is drawn toward its mode.
\model tracks the requested scale across two orders of magnitude because the explicit 3D cache supplies a per-pixel metric anchor the student cannot ignore.
Rotational errors remain comparable because rotation is not subject to this scale ambiguity.

\minisection{Robustness to camera movement severity.}
To evaluate robustness against abnormal trajectories, we synthesize ``zigzag'' camera motion by applying sinusoidal perturbations to the ground truth camera trajectory.
We define the \emph{spatial severity} $S_s$ as the oscillation amplitude relative to the total trajectory length, and the \emph{temporal severity} $S_t$ as the number of sine cycles.
As shown in \cref{fig:experiment:severity}, \model significantly reduces ATE and RTE across all noise levels, demonstrating superior robustness to severe translation jitters.
The only cell CameraCtrl II wins is RRE at the most aggressive setting.
This is likely because the implicit baseline low-pass filters its own motion, so high-frequency rotational noise is smoothed away, which lowers RRE but inflates the corresponding translation metrics.
\model instead tracks the input faithfully because the point-projection cue pulls the student toward exactly the motion requested.
The baseline learns what trajectories look like, whereas \model is geometrically constrained by them and provides more consistent metric-scale accuracy.

\minisection{Inference time.}
Measured on a single NVIDIA A100 80G with the same Wan2.2-5B video backbone, our method runs at $\sim4.05$~fps, versus $\sim0.77$~fps for CameraCtrl~II~\cite{he2025cameractrl} and $\sim0.40$~fps for Spatia~\cite{zhao2025spatia}.
This $5.3\times$ and $10.1\times$ speedup is achieved despite the additional cost of maintaining the self-refreshed 3D cache, since the cache is rebuilt only at chunk boundaries rather than per frame, while the causal streaming rollout and few-step DMD distillation jointly cut the dominant DiT cost.
This shows the potential of real-time applications for our method.

\begin{figure}[t]
	\centering
	\includegraphics[width=\linewidth]{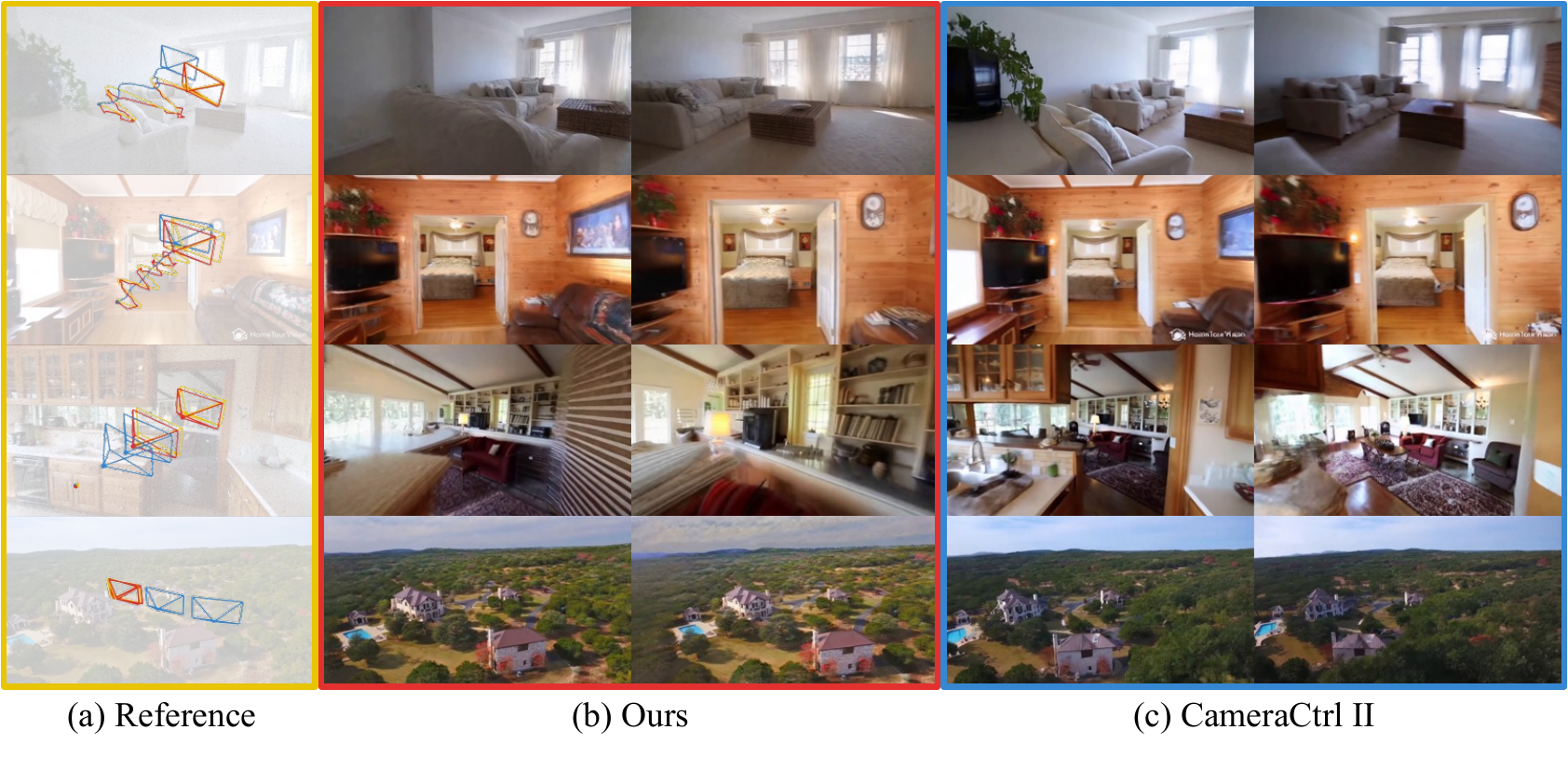}
	\caption{\textbf{Qualitative comparison of camera control ability under extreme motion.}
		1st row: spatial severity $S_s=0.4$, and temporal severity $S_t=2.0$.
		2nd row: spatial severity $S_s=0.2$, and temporal severity $S_t=4.0$.
		3rd row: magnitude factor $m=4.0$.
		4th row: magnitude factor $m=0.25$.
		Estimated trajectories from our \model (\textcolor{red}{red}) and CameraCtrl II (\textcolor{blue}{blue}) are plotted against the reference (\textcolor{yellow}{yellow}).
		CameraCtrl II oversmooths high-frequency motion (top two rows) and misestimates translation scale, leading to undershoot (3rd row) or overshoot (4th row), whereas our method tracks the input more faithfully due to explicit 3D-cache conditioning.}
	\label{fig:experiment:trajectory}
\end{figure}

\subsection{Qualitative Analysis}
\cref{fig:experiment:trajectory} overlays trajectories estimated from the generated videos of \model in \textcolor{red}{red} and CameraCtrl II in \textcolor{blue}{blue} on the reference in \textcolor{yellow}{yellow}, under the two stress tests above.
The top two rows confirm oversmoothing: the \textcolor{blue}{blue} curves collapse toward the mean while the \textcolor{red}{red} curves preserve the input jitter.
In the bottom two rows, the \textcolor{blue}{blue} curves sit at a visibly different scale, undershooting at $m\!=\!4$ and overshooting at $m\!=\!0.25$, while the \textcolor{red}{red} curves remain at the requested scale.
Our gains are therefore not an artifact of the backbone but a direct consequence of enforcing an explicit, online-refreshed 3D anchor during streaming generation.

\section{Limitation Discussion}
The two principal limitations of \model both trace back to its minimalist, online-refreshed cache design.
(i)~Memoryless cache.
We discard the previous cache at every refresh, which keeps the conditioning aligned with the current content but forgoes any long-term 3D memory.
Loop closure to a previously visited region is therefore only supported implicitly through the KV cache.
(ii)~Latency.
Each chunk runs a monocular depth estimator and a point renderer.
Real-time use will require faster depth backbones, sparser attention, or a smaller student DiT.
Despite these limitations, \model still substantially advances streaming camera control, as we summarize below in our conclusion.

\section{Conclusion}
We presented \model, a streaming video framework that couples causal autoregressive synthesis with a self-refreshed explicit 3D cache.
Our insight is that once the generator is causal, the geometric conditioning must also be causal: rebuilding the cache from the model's own most recent output closes the loop, and the resulting second-order exposure bias is handled by fully on-policy distillation, in which the point-projection conditioning during training is rendered from the student's own generated frames rather than from ground-truth geometry.
On RealEstate10K, \model substantially reduces both global and local camera-pose errors against strong recent baselines, attains state-of-the-art visual fidelity, and preserves metric-scale control far outside the training distribution, validating explicit online-refreshed geometry as a strict upgrade over both implicit and existing explicit camera conditioning.
Our on-policy geometric conditioning mechanism could, in principle, serve as a plug-in module for other streaming pipelines, and validating this generality is a promising direction for future work.

\begin{ack}
	YZ was supported in part by the SoftBank Group–ARM Fellowship.
\end{ack}

\bibliographystyle{plainnat}
\bibliography{main}

\appendix
\clearpage
\section*{Supplementary Material for \model}

\begin{figure}[t]
	\centering
	\includegraphics[width=0.7\linewidth]{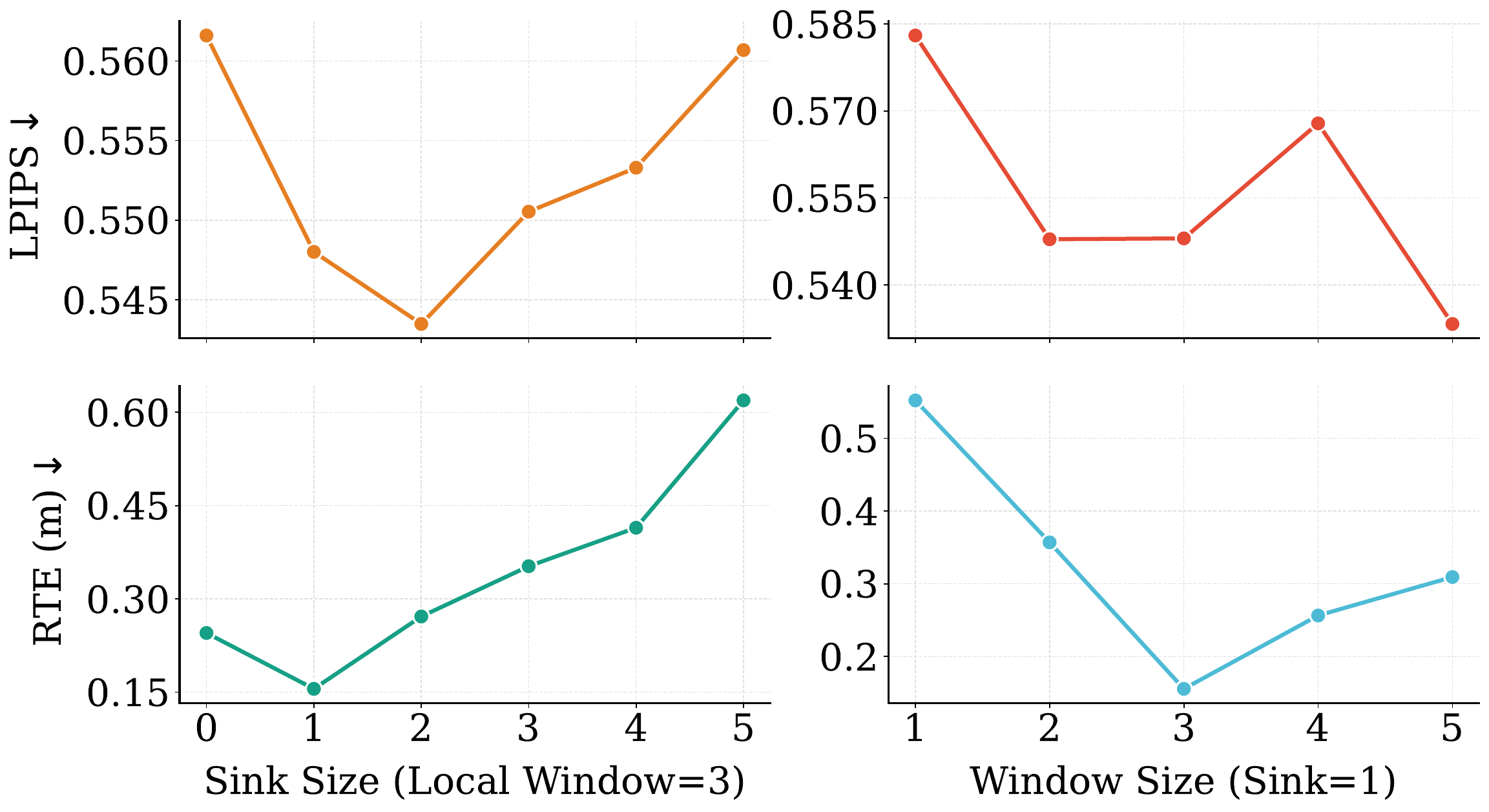}
	\caption{\textbf{Impact of sink size and local window size.}
		On 193-frame sequences, we evaluate chunk size ($c$), sink size ($s$), and local window size ($w$).
		Results show a U-shaped performance trend, with $s \in \{1, 2\}$ and $w=3$ providing the optimal balance between accuracy and efficiency.}
	\label{fig:experiment:chunk_sink_window}
\end{figure}

\begin{figure}[t]
	\centering
	\includegraphics[width=0.7\linewidth]{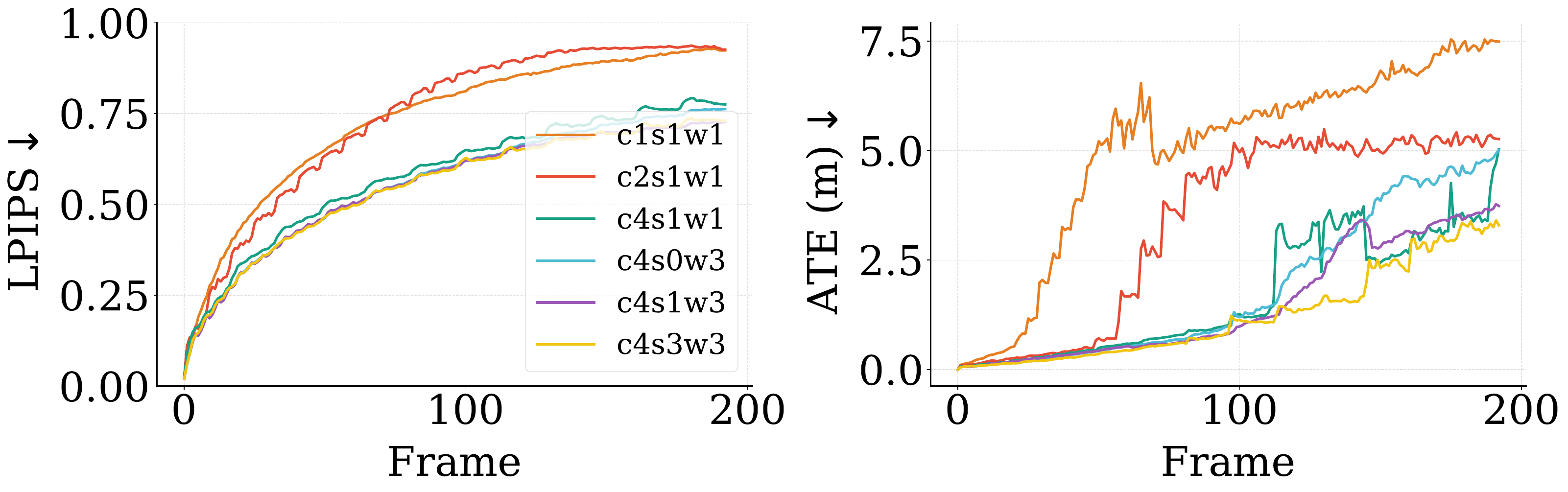}
	\caption{\textbf{Error accumulation over time.}
		Increasing chunk size ($c$) and incorporating attention sinks ($s \ge 1$) significantly stabilize long-term generation and suppress error drift compared to baseline configurations.}
	\label{fig:experiment:metric_over_time}
\end{figure}

\begin{table}[t]
	\footnotesize
	\centering
	\caption{\textbf{Ablation study on cache refreshing granularity.}
		We evaluate two temporal granularities: \textit{Every Frame} (a sliding-window update where frame $t$ is conditioned on frame $\max(0, t-c)$) and \textit{Last Frame} (a chunk-wise anchor using only the preceding chunk's last frame).
		While both modes yield comparable performance, \textit{Last Frame} refreshing offers a marginal advantage in perceptual quality, whereas \textit{Every Frame} refreshing provides a tighter geometric feedback loop, resulting in superior camera control ability.}
	\label{tab:experiment:refresh}
	\resizebox{0.85\linewidth}{!}{
		\begin{tabular}{lccccccc}
			\toprule
			Refreshing  & FVD$\downarrow$ & PSNR$\uparrow$ & SSIM$\uparrow$ & LPIPS$\downarrow$ & ATE$\downarrow$ & RTE$\downarrow$ & RRE$\downarrow$ \\
			\midrule
			Every Frame & 126.3           & 16.90          & 0.631          & 0.390             & \textbf{1.060}  & \textbf{0.116}  & 0.356           \\
			Last Frame  & \textbf{122.5}  & \textbf{17.01} & \textbf{0.648} & \textbf{0.385}    & 1.402           & 0.180           & \textbf{0.294}  \\
			\bottomrule
		\end{tabular}
	}
\end{table}

\section{Additional Details}

\minisection{Training data curation.}
To obtain paired video frames and point reprojections for training with a self-refreshed 3D cache, we estimate high-quality depth maps using MoGe-2~\cite{wang2025moge} and track camera trajectories with the off-the-shelf reconstruction method MapAnything~\cite{keetha2025mapanything}.
The ground truth camera poses and annotated depths are then aligned to the MapAnything estimates via a scale alignment procedure, ensuring that the camera control signals are at a consistent metric scale.
We then generate point reprojections that exactly follow the proposed 3D cache refresh mechanism.

\minisection{Evaluation protocol.}
We evaluate on the RealEstate10K~\cite{realestate10k} Test split by uniformly sampling 120 video clips and using them consistently for all experiments.
For our \model, we resample the videos from 30 FPS to 24 FPS.
We keep the first 81 frames for most experiments, unless otherwise specified, e.g., the long-term rollout analyses in \cref{fig:experiment:chunk_sink_window,fig:experiment:metric_over_time}.
To account for varying frame-rate requirements across different baseline methods, we resample all videos to ensure that every model operates over the same temporal span, thereby maintaining a consistent evaluation context.
At evaluation time we apply the same scale-alignment procedure used during training data curation, so that the camera control signals fed to the model remain at a consistent metric scale.

\section{Additional Quantitative Analysis}

\minisection{Chunk size, sink size, and local context window size.}
\cref{fig:experiment:chunk_sink_window} ablates the three hyperparameters $(c, s, w)$ with definition in \cref{sec:experiment:settings}.
Errors follow a U-shape in $s$ and $w$, with $s \in \{1, 2\}$ and $w=3$ best balancing drift and redundant context, while $c=4$ substantially outperforms $c=1$ or $c=2$, and additional sinks beyond $s=1$ give only marginal gains.
Intuitively, the leading chunk is the only noiseless anchor in the rollout, so retaining it as a sink preserves scene identity while extra sinks add little.
A larger $c$ aligns each step with the latent temporal granularity, letting the student denoise a coherent block in one pass.
These observations justify the default configuration $c=4$, $s=1$, $w=3$ used in all main-paper experiments.

\minisection{Long-term rollout stability.}
We further evaluate the stability of our autoregressive approach under long roll-outs on 193-frame trims of RealEstate10K, as shown in \cref{fig:experiment:metric_over_time}.
Increasing the chunk size $c$ and incorporating attention sinks ($s\ge 1$) significantly stabilize long-horizon generation and suppress error drift compared to baseline configurations, confirming that the default $(c, s, w)$ chosen above generalizes from the short-horizon regime used in the main-paper comparisons to substantially longer roll-outs.

\minisection{Granularity of 3D cache refreshing.}
We ablate the temporal granularity of 3D cache updates during inference in \cref{tab:experiment:refresh}.
\emph{Every Frame} refreshing applies a sliding-window update in which each frame $t$ is conditioned on the reprojection from frame $\max(0, t-c)$, giving a per-frame geometric anchor that tracks the rollout closely.
\emph{Last Frame} refreshing instead uses only the last frame of the preceding chunk as a single geometric anchor for the entire next chunk.
Both yield comparable performance: \emph{Last Frame} marginally improves perceptual quality by anchoring to a fixed within-chunk reference, while \emph{Every Frame} tightens the geometric feedback loop and slightly improves camera control accuracy, representing a minor trade-off between visual smoothness and geometric drift.
We adopt \emph{Every Frame} refreshing as the default in all main-paper experiments since precise metric-scale camera controllability is our primary objective.

\section{Additional Qualitative Analysis}
We refer the readers to the supplementary webpage for more video examples and qualitative analyses, including side-by-side comparisons with all baselines, stress-test sequences under varying motion magnitude and severity, and long-rollout demonstrations that highlight the temporal stability of our self-refreshed 3D cache.

\end{document}